\documentclass[10pt,twocolumn,letterpaper]{article}

\usepackage{cvpr}
\usepackage{times}
\usepackage{epsfig}
\usepackage{graphicx}
\usepackage{amsmath}
\usepackage{amssymb}
\usepackage{amsfonts}
\usepackage{subfigure}
\usepackage{multirow}
\usepackage{booktabs}
\usepackage{array}
\usepackage{color}
\usepackage{caption}
\usepackage{stfloats}
\usepackage{cite}
\usepackage{mathtools}
\usepackage{authblk}

\usepackage[table,xcdraw]{xcolor}
\usepackage[pagebackref=true,breaklinks=true,letterpaper=true,colorlinks,bookmarks=false]{hyperref}

\cvprfinalcopy 

\def\cvprPaperID{5660} 

\ifcvprfinal\fi
\begin{document}

\title{GSLAM: A General SLAM Framework and Benchmark}
\author[1]{Yong Zhao}
\author[,2]{Shibiao Xu\thanks{e-mail:shibiao.xu@nlpr.ia.ac.cn}}
\author[,1]{Shuhui Bu\thanks{e-mail:bushuhui@nwpu.edu.cn}}
\author[1]{Hongkai Jiang}
\author[1]{Pengcheng Han}

\affil[1]{Northwestern Polytechnical University}
\affil[2]{NLPR, Institute of Automation, Chinese Academy of Sciences}
\maketitle
\thispagestyle{empty}

\begin{abstract}

SLAM technology has recently seen many successes and attracted the attention of high-technological companies. However, how to unify the interface of existing or emerging algorithms, and effectively perform benchmark about the speed, robustness and portability are still problems. In this paper, we propose a novel SLAM platform named GSLAM, which not only provides evaluation functionality, but also supplies useful toolkit for researchers to quickly develop their own SLAM systems. The core contribution of GSLAM is an universal, cross-platform and full open-source SLAM interface for both research and commercial usage, which is aimed to handle interactions with input dataset, SLAM implementation, visualization and applications in an unified framework.
Through this platform, users can implement their own functions for better performance with plugin form and further boost the application to practical usage of the SLAM.

\end{abstract}


\section{Introduction}

Simultaneous Localization and Mapping (SLAM) is a hot research topic in computer vision and robotics for several decades since the 1980s \cite{durrant2006simultaneous,bailey2006simultaneous,cadena2016past}. SLAM provides fundamental function for many applications that need real-time navigation like robotics, unmanned aerial vehicles (UAVs), autonomous driving, as well as virtual and augmented reality. In recent years, SLAM technology develops rapidly and a variety of SLAM systems have been proposed, including monocular SLAM system (key-point based \cite{davison2007monoslam,klein2007parallel,mur2015orb},
direct \cite{Newcombe2011,engel2014lsd,engel2018dso} and
semi-direct methods \cite{forster2014svo,forster2017svo}), multi-sensor SLAM system (RGBD \cite{kerl2013dense,whelan2016elasticfusion,bu2016SDTAM},
Stereo \cite{engel2015sterolsd,forster2017svo,mur2017orbslam2} and inertial aided methods \cite{leutenegger2015okvis,qin2018vins,von2018vidso}), and
learning based SLAM system (supervised \cite{wang2017deepvo,brahmbhatt2017mapnet,oliveira2017topometric} and unsupervised methods\cite{zhou2017unsupervised,yin2018geonet}).

However, with the rapidly developing SLAM technology, almost all the researchers focus on the theory and implementation of their own SLAM systems, which makes it difficult to exchange ideas and not easy to port the implementation to other systems. This prevents the quick apply to various industry fields.
In addition, currently there exist many implementations of SLAM systems, how to effectively perform benchmark about the speed, robustness and portability is still a problem.
Recently, Nardi \textit{et al.} \cite{nardi2015introducing} and Bodin \textit{et al.} \cite{bodin2018slambench2} proposed uniform SLAM benchmark systems to perform quantitative, comparable and validatable experimental research for investigating trade-offs among various SLAM systems.
Through these systems, the evaluation experiments can be easily performed by using the dataset, and metric evaluation modules.

As those systems only provide evaluation benchmarks, we consider it is possible to build a platform to serve the whole life-circle of SLAM algorithms including development, evaluation and application stages.
In addition, deep learning based SLAM has achieved remarkable progress in recent years, it is necessary to create a platform which not only supports C++ but also Python for better supporting integration for geometric and deep learning based SLAM system.
Therefore, in this paper we introduce a novel SLAM platform which provides not only evaluation functionality, but also useful toolkit for researchers to quickly develop their own SLAM systems. Through this platform, frequently used functions are provided with plugin forms, therefore, users could implement their own projects with directly using them or creating their own functions for better performance.
We hope this platform could further boost the SLAM systems to practical applications.
In summary, the main contributions of this work are as follows:
\begin{enumerate}
\item We presented an universal, cross-platform and full open-source SLAM platform for both research and commercial usages, which is beyond that of previous benchmarks. The SLAM interface is consisted by several lightweight, dependency-free headers, which makes it easy to interact with different datasets, SLAM algorithms and applications with plugin forms in an unified framework. In addition, both JavaScript and Python are also provided for web based and deep learning based SLAM applications.

\item We introduced three optimized modules as utility classes including Estimator, Optimizer and Vocabulary in the proposed GSLAM platform. Estimator aims to provide a collection of close-form solvers cover all interesting cases with robust sample consensus (RANSAC); Optimizer aims to provide an unified interface for popular nonlinear SLAM problems; Vocabulary aims to provide an efficient and portable bag of words implementation for place recolonization with multi-thread and SIMD optimization.


\item Benefit from the above interface, we implemented and evaluated plugins for existing datasets, SLAM implementations and visualized applications in an unified framework, and emerging benchmark or applications could be further integrated easily in the future.
\end{enumerate}

This paper will firstly introduce the GSLAM framework interface and explain how GSLAM works.
Secondly, three utility components, Estimator, Optimizer and Vocabulary will be introduced.
Thirdly, several typical public datasets are used to evaluate different popular SLAM implementations using the GSLAM framework.
Finally, a conclusion of these works is given followed with the future works expectation.

The source code of this paper with documentation wiki can be found at: \url{https://github.com/zdzhaoyong/GSLAM}.


\section{Related Works}
In this section, we will briefly review the SLAM techniques including methods, systems and benchmarks.

\subsection{Simultaneous Localization And Mapping}
SLAM techniques build a map of an unknown environment and localize the sensor in the map with a strong focus on real-time operation.
Early SLAM are mostly based on extended kalman filter (EKF)~\cite{davison2007monoslam}. The $6$ DOF motion parameters and 3D landmarks are probabilistically represented as a single state vector. The complexity of classic EKF grows quadratically with the number of landmarks, restricting its scalability.
In recent years, SLAM technology develops rapidly and lots of monocular visual SLAM systems including key-point based \cite{davison2007monoslam,klein2007parallel,mur2015orb}, direct \cite{Newcombe2011,engel2014lsd,engel2018dso} and semi-direct methods \cite{forster2014svo,forster2017svo} are proposed.
However, monocular SLAM systems lack scale information and are not able to handle pure rotation situation, then, some other multi-sensor SLAM systems including RGBD \cite{kerl2013dense,whelan2016elasticfusion,bu2016SDTAM}, Stereo \cite{engel2015sterolsd,forster2017svo,mur2017orbslam2} and inertial aided methods \cite{leutenegger2015okvis,qin2018vins,von2018vidso} are being studied for higher robustness and precision.

While a large number of SLAM algorithms have been presented, there has little effort to unify the interface of such algorithms, or to perform a holistic comparison of their capabilities.
In addition, implementations of these SLAM algorithms are often released as standalone executables rather than as libraries, and often do not conform to any standard structure.

Recently, supervised \cite{wang2017deepvo,brahmbhatt2017mapnet,oliveira2017topometric} and unsupervised \cite{zhou2017unsupervised,yin2018geonet} deep learning based visual odometers (VO) present novel ideas compared to traditional geometry based methods,
but it is still not easy to optimize the predicted poses further for consistencies of multiple keyframes.
The tools provided by GSLAM could help them for obtaining better global consistency.
Through our framework, it is more easier to visualize or evaluate the results, and further be applied to various industry fields.


\subsection{Computer Vision and Robotics Platform}
Within the robotics and computer vision community, robotics middle-ware (e.g., ROS~\cite{QUIGLEY2009ROS}) presents a very convenient communication way between nodes and is favored by most robotics researchers.
Lots of SLAM implementations provide ROS wrapper to subscribe sensor data and publish visualization results.
But it does not unify the input and output of SLAM implementations and is hard to further evaluate different SLAM systems.

Inspired by the ROS2 \cite{maruyama2016exploring} messaging architecture, GSLAM implements a similar intra-process communication utility class named Messenger.
This provides an alternative option to replace ROS inside the SLAM implementation, and maintains the compatibility, that means all ROS defined messages are supported and ROS wrapper are naturally implemented within our framework.
Due to the intra-process design, there is no serialization and data transferring, messages are sent without latency and extra cost. Meanwhile the payloads are not limited to ROS defined messages but any copyable data structures.
Moreover, we not only provides evaluation functionality, but also supplies useful toolkit for researchers to quickly develop and integrate their own SLAM algorithm. 

\subsection{SLAM Benchmarks}
Currently, there exist several SLAM Benchmarks, including KITTI Benchmark Suite~\cite{KITTI_dataset}, TUM RGB-D Benchmarking~\cite{TUM2012STYRM} and ICL-NUIM RGB-D Benchmark Dataset~\cite{handa:etal:ICRA2014}, which only provide evaluation functionality.
In addition, SLAMBench2 \cite{bodin2018slambench2} expanded these benchmarks into algorithms and datasets, which requires users to make released implementation SLAMBench2-compatible for evaluation and it is difficult to extend to further applications.
Different from these systems, the proposed GSLAM platform provides a solution which serves the whole life-circle of the SLAM implementation from development, evaluation to application.
We provide useful toolkit for researchers to quickly develop their own SLAM system, and further visualization, evaluation and applications are developed based on an unified interface.



\section{General SLAM Framework}
The core work of GSLAM is to provide a general SLAM interface and framework. For better experience, the interface is designed to be lightweight, which is consisted by several headers and only relies on the C++11 standard library. And based on the interface, script language like JavaScript and Python are supported. In this section, the GSLAM framework is presented and a brief introduction of several basic interface classes is given.

\subsection{Framework Overview}

\begin{figure}[t]
	\centering
	\includegraphics[width=0.98\linewidth]{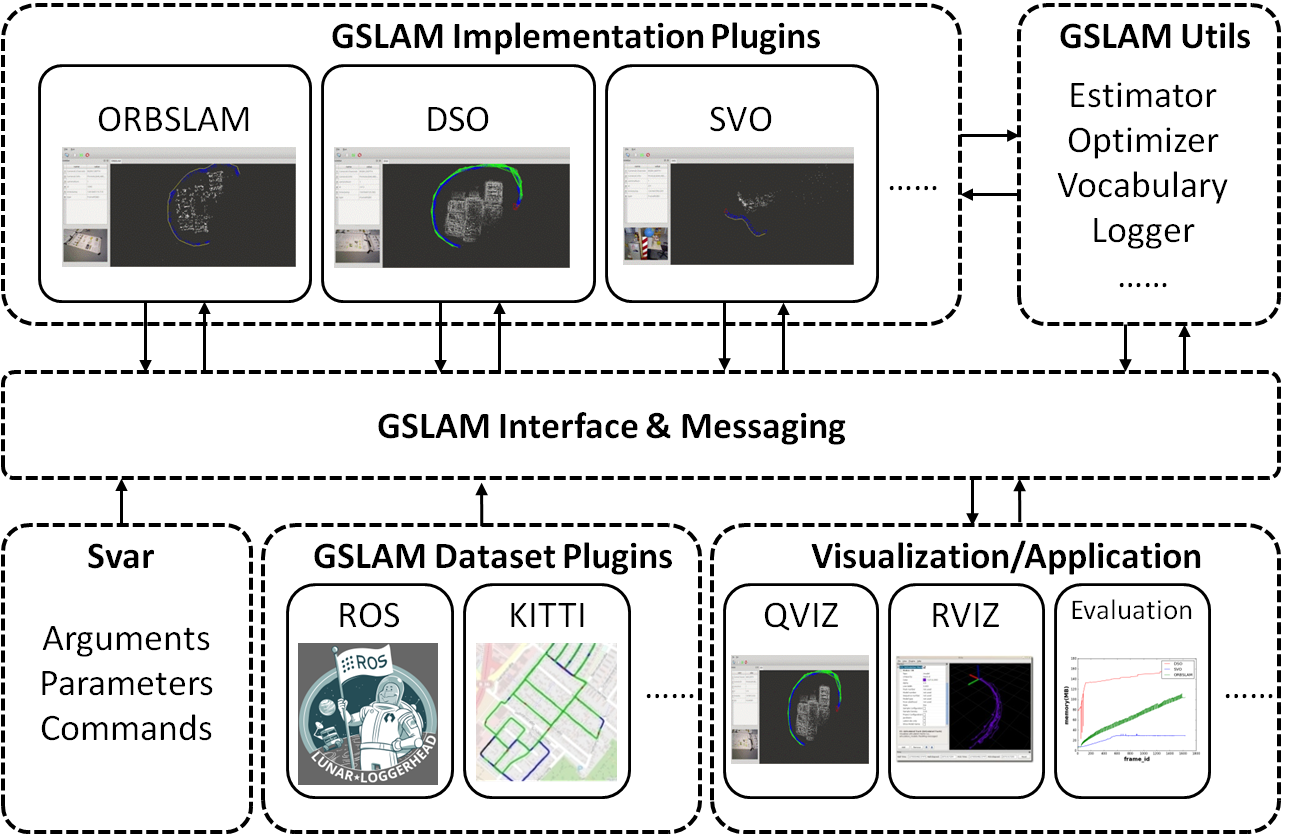}
	\caption{The framework overview of GSLAM.
	}
	\label{fig:framework}
\end{figure}

The framework of GSLAM is shown in Fig. \ref{fig:framework}, generally speaking, the interface is aimed to handle interaction with three parts:

\begin{enumerate}
\item The input of a SLAM implementation. When running a SLAM, the sensor data and some parameters are required. For GSLAM, a Svar class is used for parameters configuration and command handling. And all sensor data required by SLAM implementations are provided by a Dataset implementation and transfered using the Messenger. GSLAM implemented several popular visual SLAM datasets and users are free to implement his own dataset plugins.
\item The SLAM implementation. GSLAM treats each implementation as a plugin library. It is very easy for developers to design a SLAM implementation based on the GSLAM interface and utility classes. Developers can also wrap the implementation using the interface without extra dependency imported. Users can focus on the development of core algorithms without caring the input and output which should be handled outside the SLAM implementation.
\item The visualization part or applications using SLAM results. After SLAM implementations handled the input frames, users probably want to demonstrate or utilize the results. For generality, SLAM results should be published in a standard format. Default GSLAM uses Qt for visualization, but users are free to implement a customized visualizer and add application plugins such as an evaluation application.
\end{enumerate}

The framework is designed to be compatible with different kinds of SLAM implementations include but not limited to monocular, stereo, RGBD and multiple camera visual inertial odometer with multi-sensor fusion.
And now it best match feature based implementations while direct or deep learning based SLAM systems are also supported. As modern deep learning platforms and developers prefer Python for coding, GSLAM provides Python binding and thus developers are able to implement a SLAM using Python and call it with GSLAM or call a C++ based SLAM implementation with Python. Moreover, JavaScript is also supported for web based usages.

\subsection{Basic Interface Classes}

There are some data structures that are often used by the SLAM interface, including the parameter setting/reading, image format, pose transformation, camera model and map data structures. Here is going to give a brief introduction of some basic interface classes.

\subsubsection{Parameter Setting}
GSLAM uses a tiny arguments parsing and parameter setting class Svar, which only consists of  a single header file depending on C++11 with the following features:
\begin{enumerate}
	\item Arguments parsing and configure loading with help information. Similar to popular argument parsing tools like Google gflags \footnote{https://github.com/gflags/gflags}, the variable configuration could be loaded from arguments, files and system environment. Users could also define different types of parameters with introduction which will be shown in help.
	\item A tiny script language with variable, function and condition, which makes configure file more powerful.
	\item Thread-safe variable binding and sharing. Variables used with very high frequency are suggested to bind with pointer or reference, which provides high efficiency along with convenience.
	\item Simple function definition and calling from both C++ or plain script. A binding between command and function helps developers decouple the file dependencies.
	\item Support tree structure presentation, which means it is easy to load or save configuration with XML, JSON and YAML formats.
\end{enumerate}

\subsubsection{Intra-Process Messaging} \label{sec:messenger}
As ROS presents a very convenient communication way between nodes and is favored by most robotics researchers.
Inspired by the ROS2 messaging architecture, GSLAM implements a similar intra-process communication utility class named Messenger.
This provides an alternative option to replace ROS inside the SLAM implementation and maintains the compatibility.
Due to the intra-process design, the Messenger is able to publish and subscribe any class without extra cost. More features are listed below:

\begin{enumerate}
	\item The interface keeps ROS style and easily for users to get started. And all ROS defined messages are supported, which means very few works are needed to replace the original ROS messaging.
	\item Since there is no serialization and data transferring, messages can be sent without latency and extra cost. Meanwhile the payload is not limited to ROS defined messages but any copyable data structures are supported.
    \item The source are header files only based on C++11 with no extra dependency, which makes it portable.
	\item The API is thread-safe and supports multi-thread condition notify when the queue size is greater than zero. Both topic name and RTTI data structure check are done before a publisher and subscriber are connected from each other to ensure correct calls.
\end{enumerate}

\subsubsection{3D Transformation}
Rotation, rigid and similarity are three of the most used transformations in SLAM research.
A similarity transformation of a point $\mathbf{p}=(x,y,z)^T$ is common to use a $4 \times 4$ homogeneous transformation matrix or decompose such a matrix into rotational and translational components:

\begin{equation}
\left[\begin{array}{c}\mathbf{p}' \\ 1\end{array} \right] =
\left[\begin{array}{cc}s\mathbf{R}& \mathbf{t} \\
\mathbf{0}^T& 1 \end{array} \right]
\left[\begin{array}{c}\mathbf{p} \\ 1\end{array} \right].
\end{equation}

Here $\mathbf{R} \in \mathbb{R}^{3\times3}$ represents the rotation matrix, which is given as a member of the $SO(3)$ Lie group \cite{selig2004lie} with three unit direction axises. $\mathbf{t \in \mathbb{R}^3}$ means the translation and $s$ is the scale factor. The similarity transform matrix belongs to the $SIM(3)$ group. When the scale $s=1$, the transform becomes a rigid transform and belongs to the $SE(3)$ group.

\begin{table}[tb]
	\caption{Transform comparison with three popular implementations. The table statistics the time usage to run 1e6 times of transform multiply, point transform, exponential and logarithm in Milli seconds on an i7-6700 CPU running 64bit Ubuntu.} \vspace{-5mm}
	\label{tab:transform}
	\begin{center}
		\begin{tabular}{c|c|cccc}
			\hline
			\multicolumn{2}{c|}{Method}   	
							& GSLAM    			& Sophus  & TooN  & Ceres \\
			\hline
			\multirow{4}{*}{$SO(3)$}
			&\textit{mult}  & \textbf{14.9}     & 34.3    & 17.8 	&159.1 \\
			&\textit{trans}	& 15.4    			& 17.2    & \textbf{14.5} 	&90.4 	 \\
			&\textit{exp}  	& \textbf{80.7}     & 98.4    & 106.8 	&- \\
			&\textit{log}   & \textbf{55.7}     & 72.5    & 63.8 	&- \\
			\cline{0-1}
			\multirow{4}{*}{$SE(3)$}
			&\textit{mult}	& \textbf{28.6}    	& 55.2    & 29.3 	&- \\
			&\textit{trans} & 19.3    			& 19.8    & \textbf{12.1} 	&- \\
			&\textit{exp}	& 152.4    			& 249.2   & \textbf{99.2} 	&-\\
			&\textit{log}	& \textbf{152.7}	& 194.0   & 205.8 	&- \\
			\cline{0-1}
			\multirow{4}{*}{$SIM(3)$}
			&\textit{mult}	& \textbf{33.2}		& 58.5    & 34.5 	&-  \\
			&\textit{trans} & 16.9    			& 17.2    & \textbf{13.7} 	&-  \\
			&\textit{exp}	& \textbf{180.2}    & 286.8   & 229.0 	&-  \\
			&\textit{log}	& \textbf{202.5} 	& 341.6   & 303.6 	&-  \\
			\hline
		\end{tabular}
	\end{center}
\end{table}

For the rotational component, there are several choices for representation, including the matrix, Euler angle, unit quaternion and Lie algebra $so(3)$. For a given transformation, we can use any of these for representation and can convert one to another. However, we need to pay close attention to the selected representation when we consider multiple transformations and manifold optimization. The matrix representation is overparamatrized with 9 parameters where as the rotation only has 3 degrees of freedom (DOF). The Euler angle representation uses 3 variable and is easy to understand but faces the well-known “gimbal lock” problem and not convenience to multiple transformations. The unit quaternion is the most efficient way to perform multiple and Lie algebra is the common representation to perform manifold optimization. The matrix representation of rotation $\mathbf{R}$ is calculated from $\phi \in \mathbb{R}^3$ using the exponential function according to Lie algebra $so(3)$:
\begin{eqnarray}
\mathbf{R}&=&\exp(\phi^{\wedge})=\exp(\theta \mathbf{a}^{\wedge})\\
&=&\cos \theta \mathbf{I}+(1-\cos \theta)\mathbf{a}\mathbf{a}^T+\sin \theta\mathbf{a}^T.
\end{eqnarray}
Where $\mathbf{a}$ is the rotation axis and $\theta$ is the angle to rotate. $\phi^{\wedge}$ is the skew-symmetric matrix of $\phi$.

Similarly the Lie algebra of rigid and similarity transformation $se(3)$ and $sim(3)$ are defined. GSLAM uses quaternion to represent the rotational component and provide functions converting from one representation to other representations. Table \ref{tab:transform} demonstrates our transforms implementation with comparison to three other popular implementations Sophus, TooN and Ceres.
Since Ceres implementation uses the angle axis representation, the rotation exponential and logarithm are not needed. As the table demonstrates, the GSLAM implementation outperforms due to the use of quaternion and better optimization, while TooN utilizes the matrix implementation and outperforms on point transformation.

%

\subsubsection{Image Format}

Image data storing and transferring are two of the most important functions for visual SLAM. For efficiency and convenience, GSLAM utilizes a data structure GImage which is compatible to cv::Mat. It has a smart point counter for safely memory free and is easy to be transfered without memory copy. And the data pointer is aligned so that it would be easier for single instruction multiple data (SIMD) speed up. Users can convert between GImage and cv::Mat seamlessly and safely without memory copy.

\subsubsection{Camera Models}

A camera model should be defined to project a 3D point $\mathbf{p}_c$ from camera coordinates to 2D pixel $\mathbf{x}$. One most popular camera model is the pinhole model where the projection can be represented by multiply an intrinsic matrix $K$ known as:
\begin{equation}
\mathbf{x}=\mathbf{K}\mathbf{p}_c
=\left[\begin{array}{ccc} f_x& &c_x\\ &f_y&c_y\\& & 1 \end{array} \right]\mathbf{p}_c
\end{equation}

As images for SLAM possibly contain radial and tangential distortion due to imperfect manufacturing or are captured with a fish-eye or panorama camera, different camera models are proposed to describe the projection. GSLAM provides implementations including the OpenCV \cite{fryer1986lens} (used by ORBSLAM \cite{mur2017orbslam2}), ATAN (used by PTAM \cite{klein2007parallel}) and OCamCalib \cite{scaramuzza2006flexible} (used by MultiCol-SLAM \cite{urban2016multicol} ).
Users are also easy to inherit the class and implement some other camera models like Kannala-Brandt \cite{kannala2006generic} and Equirectangular panorama model.

\subsubsection{Map Data Structure}\label{sec:mapdatastructure}
For a SLAM implementation, its goal is to localize the real-time poses and generate a map.
GSLAM suggests an unified map data structure which is consisted by several mapframes and mappoints.
This data structure is appropriate for most of the existed visual SLAM systems including both feature based or direct methods.

Mapframes are used to represent location statuses in different times with various information captured by sensors or estimated results including IMU or GPS raw data, depth information and camera models.
Relationships between them are estimated by SLAM implementations and their connections form a pose graph.

Mappoints are used to express the environment observed by frames, which are generally used by feature based methods.
However, a mappoint could not only represents a key-point but also a GCP, edge line or 3D object.
Their correspondences with mapframes form an observation graph which are often called as bundle graph.

\begin{table}[tb]
	\caption{Algorithms which the GSLAM Estimator implemented.} \vspace{-5mm}
	\label{tab:estimator}
	\begin{center}
		\begin{tabular}{c|c|cc}
			\hline
			\multicolumn{2}{c|}{Algorithm}   	
									& Ref.    						& Model  	 \\
			\hline
			\multirow{7}{*}{2D-2D}
			&\textit{F8-Point}  	& \cite{fischler2014readings}  	& Fundamental    \\
			&\textit{F7-Point}  	& \cite{hartley2003multiple}  	& Fundamental    \\
			&\textit{E5-Stewenius}  & \cite{stewenius2006recent}	& Essential    	 \\
			&\textit{E5-Nister}  	& \cite{nister2004efficient}	& Essential    	 \\
			&\textit{E5-Kneip}  	& \cite{kneip2012finding}  		& Essential    	 \\
			&\textit{H4-Point}  	& \cite{hartley2003multiple}  	& Homography   	 \\
			&\textit{A3-Point}  	& \cite{bradski2000opencv}  	& Affine2D    	 \\
			\hline
			\multirow{6}{*}{2D-3D}
			&\textit{P4-EPnP}  		& \cite{lepetit2009epnp}  		& SE3  	 		\\
			&\textit{P3-Gao}  		& \cite{gao2003complete}		& SE3      		\\
			&\textit{P3-Kneip}  	& \cite{kneip2011novel}			& SE3     		\\
			&\textit{P3-GPnP}  		& \cite{kneip2013using}			& SE3     		\\
			&\textit{P2-Kneip}  	& \cite{kneip2011robust}		& SE3      		\\
			&\textit{T2-Triangulate}& \cite{kneip2014opengv}		& Translation   \\
			\hline
			\multirow{3}{*}{3D-3D}
			&\textit{A4-Point} 		& \cite{bradski2000opencv}		& Affine3D     \\
			&\textit{S3-Horn}  		& \cite{horn1987closed}  		& SIM3  	   \\
			&\textit{P3-Plane}  	& \cite{kneip2011novel}			& SE3          \\
			\hline
		\end{tabular}
	\end{center}
\end{table}


\section{SLAM Implementation Utilities}
For making things easier to implement a SLAM system, GSLAM provides some utility classes.
This section will briefly introduce three optimized modules named Estimator, Optimizer and Vocabulary.

\subsection{Estimator}
The purely geometric computation remains a fundamental problem that requires robust and accurate real-time solutions.
Both classical visual SLAM algorithms \cite{klein2007parallel,forster2014svo,mur2015orb}) or modern visual-inertial solutions  \cite{leutenegger2015okvis,qin2018vins,von2018vidso} rely on geometric vision algorithms for initialization, relocalization and loop-closure.
OpenCV \cite{bradski2000opencv} provides several geometry algorithms and Kneip presents a toolbox for geometric vision OpenGV \cite{kneip2014opengv} which is limited to camera pose computation.
Estimator of GSLAM aims to provide a collection of close-form solvers cover all interesting cases with robust sample consensus (RANSAC) \cite{fischler1981random} methods.

Table \ref{tab:estimator} lists the algorithms supported by the estimator.
They are divided into three categories according to the given observations.
2D-2D matches are used to estimate epipolar or homography constraints and relative pose could be decomposed from them.
2D-3D correspondences are used to estimate both central or non-central absolute pose for monocular or multiple camera systems, which is the famous PnP problem.
3D geometry functions such as plane fitting, and estimating the $SIM3$ transformation of two point clouds are also supported.
Most algorithms are implemented depending on the open-source linear algebra library Eigen, which is header-only and readily for most platforms.

\subsection{Optimizer}

Nonlinear optimization is the core part of state-of-the-art geometric SLAM systems.
Due to the high dimensionality and sparseness of Hessian matrix, graph structures are used to modeling complex estimation problems for SLAM.
Several frameworks including Ceres \cite{ceres-solver}, G2O \cite{grisetti2011g2o} and GTSAM \cite{dellaert2012gtsam} are proposed to solve general graph optimization problems.
These frameworks are popular used by different SLAM systems.
ORB-SLAM \cite{mur2015orb,mur2017orbslam2}, SVO \cite{forster2014svo,forster2017svo} use G2O for bundle adjustment and pose graph optimization.
OKVIS \cite{leutenegger2015okvis}, VINS \cite{qin2018vins} use Ceres for graph optimization with IMU factors and sliding window is used to control the computation complex.
Forster \textit{et al.} present a visual-initial method \cite{forster2017manifold} based on SVO and implement the back-end with GTSAM.

Optimizer of GSLAM aims to provide an unified interface for most of nonlinear SLAM problems such as PnP solver, bundle adjustment, pose graph optimization.
A general implementation plugin for these problems is carried out based on the Ceres library.
For a particular problem such as bundle adjustment, some more efficient implementations such as PBA \cite{wu2011pba} and ICE-BA \cite{liu2018cvpr} could also be provided as a plugin.
With the optimizer utility, developers are able to access different implementations with an united interface, particularly for deep learning based SLAM systems.

\setlength{\tabcolsep}{3pt}
\begin{table}[tb]
	\caption{Comparison of four BoW implementations in loading, saving and training a same vocabulary with memory usage statistics. The experiment is performed on a computer with i7-6700 CPU, 16GB RAM running 64bit Ubuntu. 400 and 10k images from DroneMap \cite{bu2016map2dfusion} dataset are used to train the models with 4 and 6 levels.} \vspace{-5mm}
	\label{tab:vocabulary}
	\begin{center}
		\begin{tabular}{c|c|cccc}
			\hline
			\multicolumn{2}{c|}{Implementation}   	
								& Ours    			& DBoW2  	& DBoW3  	& FBoW \\
			\hline
			\multirow{4}{*}{Load}
			&\textit{ORB-4}  	& \textbf{67.3us}  	& 47.2ms    & 7.1ms 	&72.3us \\ 
			&\textit{ORB-6}  	& \textbf{7.2ms}  	& 6.8 s    	& 1.1 s 	&9.5ms \\
			&\textit{SIFT-4}	& \textbf{1.0ms}   	& 436.1ms   & 5.1ms 	&1.1ms\\
			\cline{0-1}
			\multirow{4}{*}{Save}
			&\textit{ORB-4}		& \textbf{437.9us}  & 40.4ms    & 1.7ms 	&553.1us \\
			&\textit{ORB-6} 	& 34.4ms   			& 4.8 s    	& 632.4ms 	&\textbf{20.6ms} \\
			&\textit{SIFT-4}	& 4.4ms				& 437.6ms   & 6.7ms 	&\textbf{2.7ms}\\
			\cline{0-1}
			\multirow{4}{*}{Train}
			&\textit{ORB-4}		& \textbf{7.6 s}	& 24.8 s    & 23.6 s 	&8.5 s  \\
			&\textit{ORB-6} 	& \textbf{230.5 s}  & 1.1Ks     & 911.4 s 	&270.4 s  \\
			&\textit{SIFT-4}	& 23.5 s    		& 327.7 s   & 299.0 s 	&\textbf{18.7 s}  \\
			\cline{0-1}
			\multirow{4}{*}{Trans}
			&\textit{ORB-4}		& \textbf{615.5us}	& 2.1ms   	& 1.9ms 	&862.4us  \\
			&\textit{ORB-6} 	& \textbf{723.7us}	& 6.0ms    	& 4.9ms 	&1.2ms  \\
			&\textit{SIFT-4}	& \textbf{1.1ms}    & 10.3ms   	& 9.2ms 	&11.5ms  \\
			\cline{0-1}
			\multirow{4}{*}{Mem}
			&\textit{ORB-4}		& \textbf{0.44MB}	& 2.5MB  	& 2.5MB 	&0.45MB  \\
			&\textit{ORB-6} 	& \textbf{44.4MB}   & 247.1MB   & 246.5MB 	&45.3MB  \\
			&\textit{SIFT-4}	& \textbf{5.8MB}    & 7.8MB   	& 7.8MB 	&5.8MB  \\
			\hline
		\end{tabular}
	\end{center}
\end{table}

\subsection{Vocabulary}
Place recognition is one of the most important part for SLAM relocalization and loop detection.
Bag of words (BoW) approach is popular used in SLAM systems since its efficiency and performance.
FabMap \cite{cummins2008fabmap} \cite{glover2012fabmap} propose a probabilistic approach to the problem of recognizing places based on their appearance, which is used by RSLAM \cite{mei2011rslam}, LSD-SLAM \cite{engel2014lsd}.
As it uses float descriptors like SIFT and SURF, DBoW2 \cite{galvez2012dbow2} builds a vocabulary tree for training and detection, which supports both binary and float descriptors.
Rafael presents two improved versions of DBoW2 named DBoW3 and FBoW \cite{rmsalinas2017fbow}, which simplify the interface and accelerate the training and loading speed.
After ORB-SLAM \cite{mur2015orb} adopts the ORB \cite{ORB_2011} descriptor and uses DBoW2 for loop detection \cite{mur2014fast}, relocalization and fast matching, a varies of SLAM systems such as ORB-SLAM2 \cite{mur2017orbslam2}, VINS-Mono \cite{qin2018vins} and LDSO \cite{gao2018ldso} use DBoW3 for loop detection. It has become the most popular tool to implement place recognition for SLAM systems.

Inspired by the above works, GSLAM carries out a header only implementation of DBoW3 vocabulary with the following features:

\begin{enumerate}
	\item Removed OpenCV dependency and the all functions are implemented within one single header file only depending on C++11.
	\item Combined the advantages of both DBoW2/3 and FBoW \cite{rmsalinas2017fbow} which are extremely fast and easy to use. Interface similar to DBoW3 is provided and both binary and float descriptors are accelerated using SSE and AVX instructions.
	\item We improved the memory usage and accelerated the speed of loading, saving or training a vocabulary and transformation from features of images to a BoW vector.
\end{enumerate}

A comparison of the four implementations is demonstrated in Table \ref{tab:vocabulary}. In the experiment, each parent node has 10 children, and for ORB feature detection we use the ORB-SLAM \cite{mur2017orbslam2}, and SiftGPU \cite{wu2011siftgpu} is used for SIFT detection. Two ORB vocabularies with 4 and 6 levels and one SIFT vocabulary are used in the results. Both FBoW and GSLAM use multi-thread for vocabulary training. Our implementation outperforms in almost all items including loading and saving the vocabulary, training a new vocabulary, transforming a descriptor list to a BoW vector for place recognition and a feature vector for fast feature matching.
Furthermore, the GSLAM implementation uses less memory and allocates less pieces of dynamic memories as we found that the fragmentation problem is the main reason that DBoW2 needs lots of memory.


\section{SLAM Evaluation Benchmark}

Existed benchmarks \cite{geiger2012we,sturm2012benchmark} need users download test datasets and upload results for precision evaluation, which are not able to unify the running environment and evaluate a fairy performance comparison.
Benefit from the unified interface of GSLAM, the evaluation of SLAM systems becomes much more elegant. With help of GSLAM, developers just need to upload the SLAM plugin and various evaluations on speed, computation cost and accuracy could be done in a dockerlized environment with fixed resources.

\begin{table}[tb]
	\caption{Some open dataset plugins build-in implemented by now.} \vspace{-5mm}
	\label{tab:datasets}
	\begin{center}
		\begin{tabular}{cccccc}
			\hline
			Dataset    								& Year   & Environment  & Type 	    \\
			\hline	
			KITTI\cite{geiger2012we}				& 2012	 & outdoors 	& multi-cam, imu  \\
			TUMRGBD\cite{sturm2012benchmark} 	    & 2012   & indoors   	& RGBD  \\
			ICL \cite{handa2014benchmark} 	    	& 2014   & simulation  	& RGBD  \\
			TUMMono \cite{engel2016photometrically}	& 2016   & indoors   	& mono  \\
			Euroc \cite{burri2016euroc}				& 2016   & indoors 		& stereo, imu  \\
			NPUDroneMap \cite{bu2016map2dfusion}	& 2016   & aerial  		& mono  \\
			TUMVI \cite{schubert2018tum}			& 2018   & in/outdoors  & stereo, imu  \\
			CVMono \cite{bradski2000opencv}			&  -   	 & -  			& mono  \\
			ROS \cite{QUIGLEY2009ROS}				&  -   	 & -  			& -  \\
			\hline
		\end{tabular}
	\end{center}
\end{table}

In this section we will firstly introduce some datasets and SLAM plugins implemented.
And then an evaluation is carried out with three representative SLAM implementations on speed, accuracy, memory and CPU usages.
The evaluation is performed to demonstrate the possibility of an united SLAM benchmark with different SLAM implementation plugins.

\subsection{Datasets}

A sensor data stream with corresponding configuration is always needed to run a SLAM system.
For letting developers focus on the development of the core SLAM plugins, GSLAM provides a standard dataset interface where developers do not need to take care of the SLAM inputs.
Both online sensor input and offline data are provided through different dataset plugins, and correct plugin is dynamic loaded by identify the given dataset path suffix.
A dataset implementation should provide all sensor stream requested with related configurations, thus no extra setting is needed for different datasets.
All different sensor streams are published through Messenger introduced in Sec. \ref{sec:messenger} with standard topic names and data formats.

GSLAM has already implemented several popular visual SLAM dataset plugins which are listed in Table. \ref{tab:datasets}.
It is also very easy for users to implement a dataset plugin based on the header-only GSLAM core and publish it as a plugin or compile it along with the applications.

\subsection{SLAM Implementations}

\begin{figure}[t]
	\centering
	\includegraphics[width=0.49\linewidth]{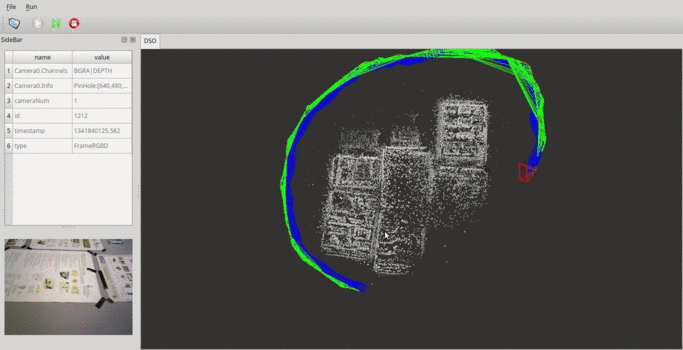}
	\includegraphics[width=0.49\linewidth]{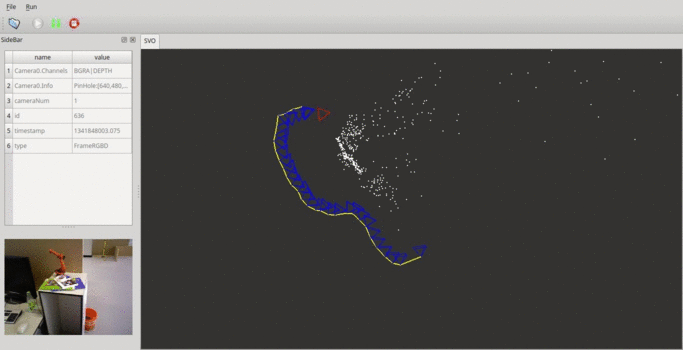}
	\includegraphics[width=0.49\linewidth]{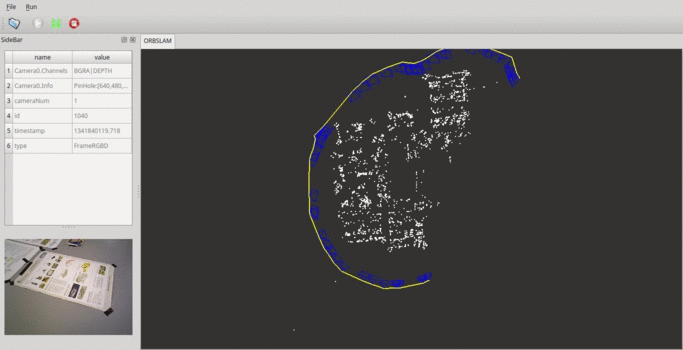}
	\includegraphics[width=0.49\linewidth]{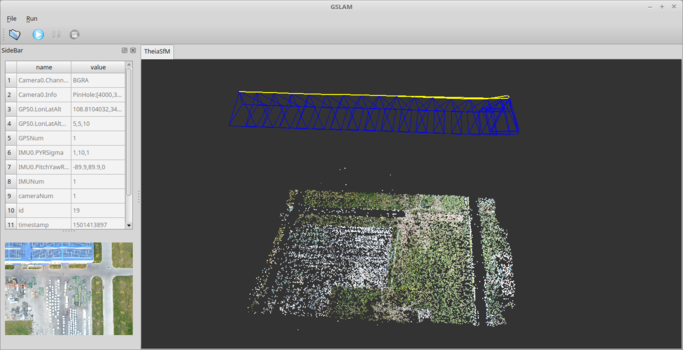}
	\caption{Screenshots of some SLAM and SfM plugins implemented, including direct method DSO \cite{engel2018dso} (left-top), semi-direct visual odometer SVO \cite{forster2014svo,forster2017svo} (right-top), feature based method ORBSLAM \cite{mur2015orb,mur2017orbslam2} (left-bottom) and global SfM system TheiaSfM \cite{sweeney2015theia} (right-bottom). }
	\label{fig:slams}
\end{figure}

Fig.\ref{fig:slams} demonstrates the screenshots of some open-source SLAM and SfM plugins running with build-in Qt visualizer. Different architectures of SLAM systems including direct, semi-direct, feature based and even SfM methods are supported by our framework.
It should be mentioned that since SVO, ORBSLAM and TheiaSfM utilize the map data structure introduced in Sec. \ref{sec:mapdatastructure}, the visualization is auto supported.
The DSO implementation needs to publish the results such as pointcloud, camera poses, trajectory and pose graph for visualization just like the ROS based implementation does.
Users are able to access different SLAM plugins with the unified framework and it is very convenient to develop a SLAM based applications depending on the C++, Python and Node-JS interfaces.
As many researchers use ROS for development, GSLAM also provides the ROS visualizer plugin to transfer the ROS defined messages seamlessly, and developers could utilize Rviz for display or continue to develop other ROS based applications.

\begin{figure*}[t]
	\centering	
	\subfigure[\textit{Memory usage}]{
		\includegraphics[width=0.236\textwidth]{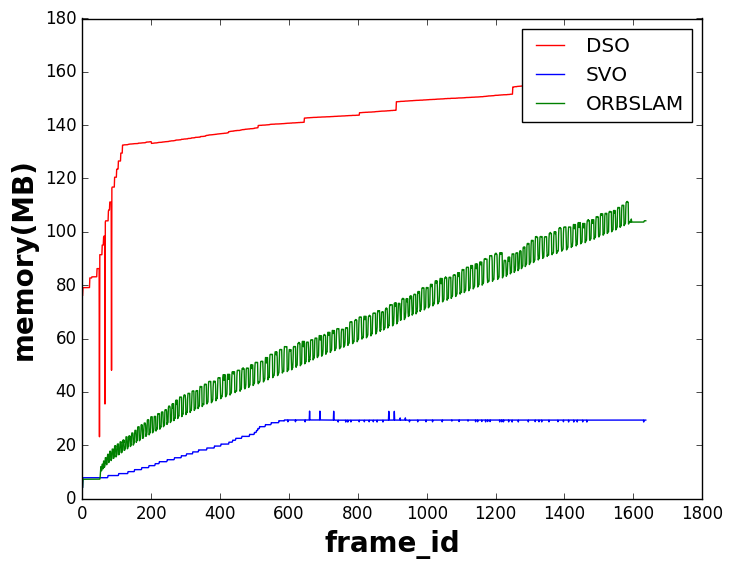}
	}
	\subfigure[\textit{Memory malloc number}]{
		\includegraphics[width=0.236\textwidth]{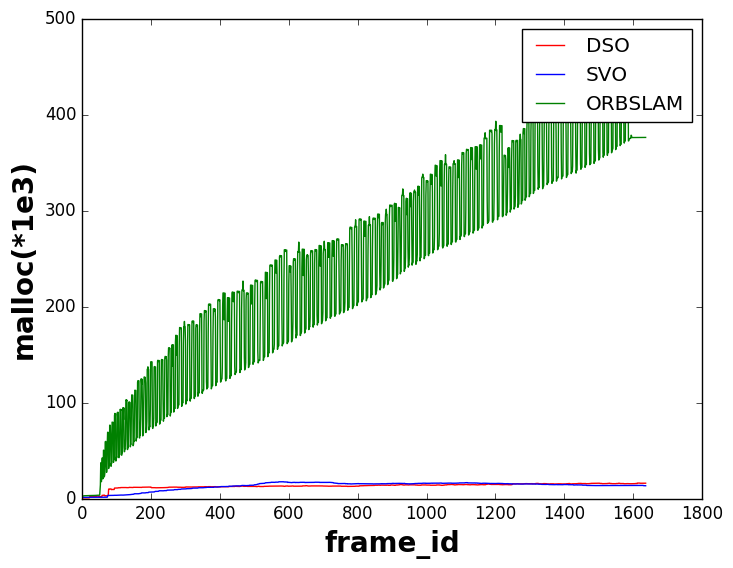}
	}
	\subfigure[\textit{CPU usage}]{
		\includegraphics[width=0.236\textwidth]{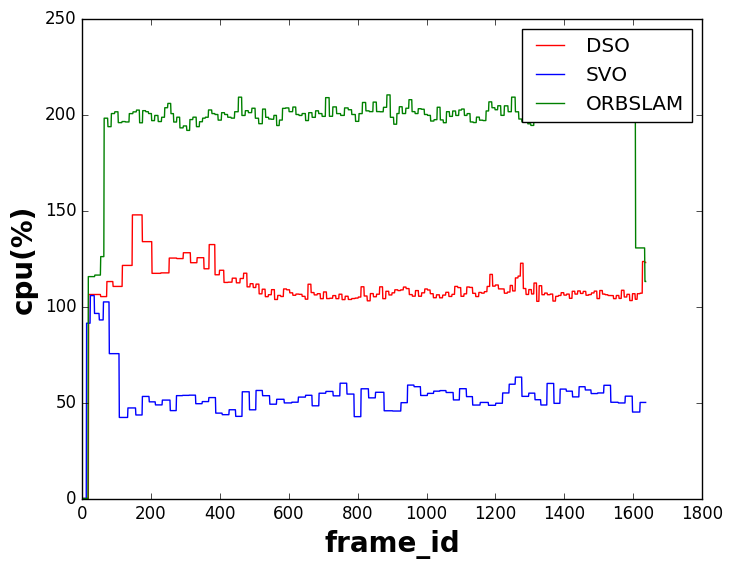}
	}
	\subfigure[\textit{Frame duration}]{
		\includegraphics[width=0.236\textwidth]{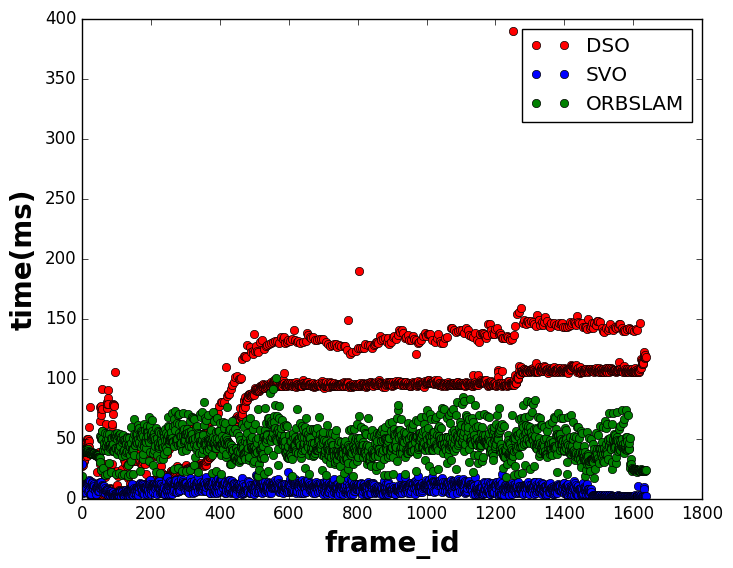}
	}
	\caption{Computation performance statistics of three monocular implementations integrated within the evaluation tool. The recordings of memory usage and memory allocated numbers are started after the SLAM application loaded, and updated after every frame processed. CPU usage is updated when the process occupied CPU time increases a curtain value. Frame duration is measured by the time between current frame published and processed.
		}
	\label{fig:computation}
\end{figure*}

\begin{figure*}[t]
	\centering
	\subfigure[\textit{Trajectory aligned}]{\label{fig:trajectory}
		\includegraphics[width=0.23\textwidth]{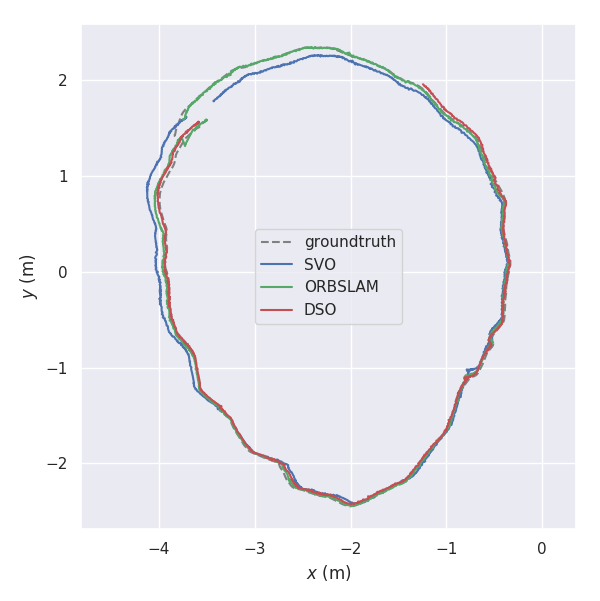}
	}
	\subfigure[\textit{DSO}]{
		\includegraphics[width=0.23\textwidth]{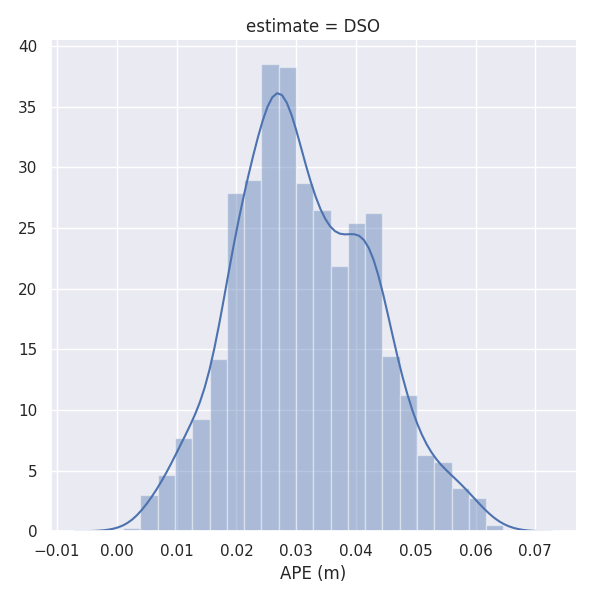}
	}
	\subfigure[\textit{SVO}]{
		\includegraphics[width=0.23\textwidth]{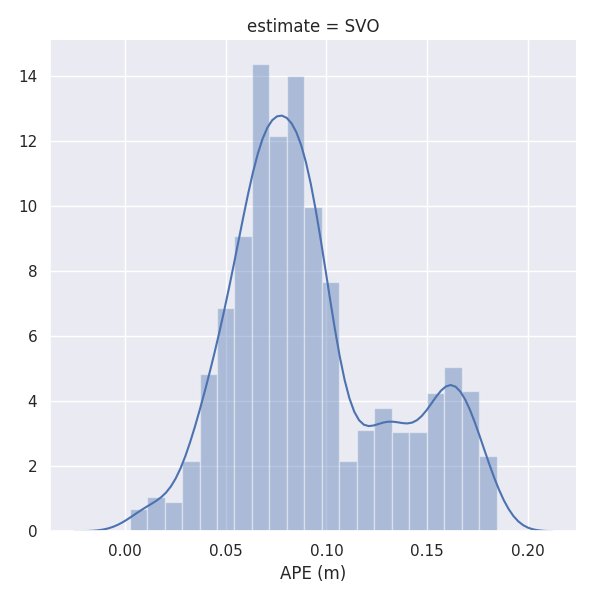}
	}
	\subfigure[\textit{ORBSLAM}]{
		\includegraphics[width=0.23\textwidth]{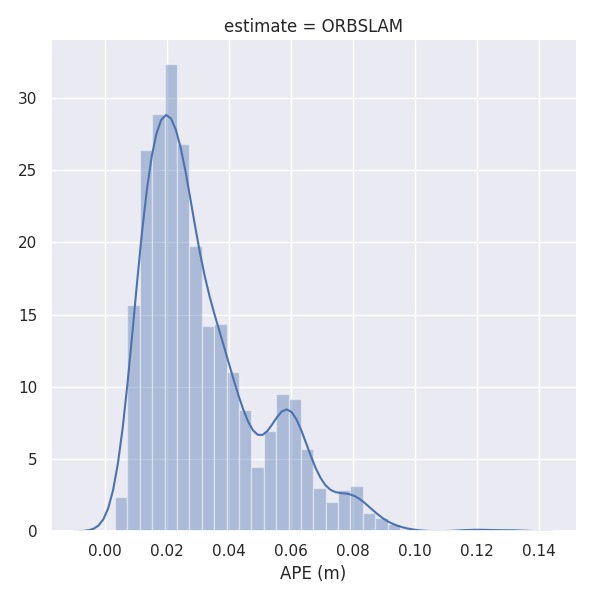}
	}
	\caption{Odometer trajectories aligned with ground-truth (left) and absolute pose error (APE) distributions of DSO, SVO and ORBSLAM. The odometer trajectories published lively instead of final results are used.
	}
	\label{fig:precision}
\end{figure*}

\subsection{Evaluation}

As most existing benchmarks only provide datasets with or without ground-truth for users to perform evaluations by themselves.
GSLAM provides a build-in plugin and some script tools for both computation performance and accuracy evaluation.

The sequence \textit{nostructure-texture-near-withloop} from TUMRGBD dataset is used to demonstrate how the evaluation performs.
And three open-source monocular SLAM plugins DSO, SVO and ORBSLAM are adopted for the following experiments.
%
A computer with i7-6700 CPU, GTX 1060 GPU and 16GB RAM running 64bit Ubuntu 16.04 is used for all the experiments.



The computation performance evaluation including memory usage, malloc numbers, CPU usage and time used by every-frame statistics are shown in Fig. \ref{fig:computation}.
The results demonstrate that SVO uses the least memory, CPU resources and obtains fastest speed.
And all cost keeps stable since SVO is a visual odometer and just a local map is maintained inside the implementation.
DSO mallocs fewer memory block numbers, but consumes more than 100MB RAM which increases slowly.
One problem of DSO is that the processing time increases dramatically when frame number is below 500, in addition, the processing times for key-frames are even longer.
ORBSLAM uses the most CPU resources and the computation time is stable, but the memory usage increases fast and it allocates and frees a lot of memory blocks since the bundle adjustment uses the G2O library and no incremental optimization approach is used.


The odometer trajectory evaluation is presented in Fig. \ref{fig:precision}. As we can see, SVO is faster but have much higher drift, while ORBSLAM achieves the best accuracy in terms of absolute pose error (APE).
%
%
The relative pose error (RPE) are also provided, however due to the limitation of the paragraph, more experimental results are provided in the supplementary materials.
Since the integrated evaluation is a pluggable plugin application, it can be reimplemented with more evaluation metrics such as the precision of pointcloud.


\section{Conclusions}
In this paper, we introduce a novel and generic SLAM platform named GSLAM, which provides support from development, evaluation to application.
Through this platform, frequently used toolkits are provided by plugin form, and users can also easily develop their own modules.
To make the platform easy to be used, we make the interfaces only depend C++11.
In addition, Python and JavaScript interfaces are provided for better integrating traditional and deep learning based SLAM or distributed on the web.

%
We hope that researchers and engineers will find GSLAM is an useful platform for practical development, and it could further boost the applications of SLAM to various fields.
In the following research, more SLAM implementations, documents and demonstrate code will be supplied for easy to learn and use. In addition, integration of traditional and deep learning based SLAM will be provided to further explore the undiscovered possibilities of SLAM systems.



{\small
	\bibliographystyle{ieee}
	\bibliography{egbib}

\begin{thebibliography}{10}\itemsep=-1pt

\bibitem{ceres-solver}
S.~Agarwal, K.~Mierle, and Others.
\newblock Ceres solver.
\newblock \url{http://ceres-solver.org}.

\bibitem{bailey2006simultaneous}
T.~Bailey and H.~Durrant-Whyte.
\newblock Simultaneous localization and mapping (slam): Part ii.
\newblock {\em IEEE Robotics \& Automation Magazine}, 13(3):108--117, 2006.

\bibitem{bodin2018slambench2}
B.~Bodin, H.~Wagstaff, S.~Saecdi, L.~Nardi, E.~Vespa, J.~Mawer, A.~Nisbet,
  M.~Luj{\'a}n, S.~Furber, A.~J. Davison, et~al.
\newblock Slambench2: Multi-objective head-to-head benchmarking for visual
  slam.
\newblock In {\em 2018 IEEE International Conference on Robotics and Automation
  (ICRA)}, pages 1--8. IEEE, 2018.

\bibitem{bradski2000opencv}
G.~Bradski et~al.
\newblock The opencv library (2000).
\newblock {\em Dr. Dobb’s Journal of Software Tools}, 2000.

\bibitem{brahmbhatt2017mapnet}
S.~Brahmbhatt, J.~Gu, K.~Kim, J.~Hays, and J.~Kautz.
\newblock Mapnet: Geometry-aware learning of maps for camera localization.
\newblock {\em arXiv preprint arXiv:1712.03342}, 2017.

\bibitem{bu2016SDTAM}
S.~Bu, Y.~Zhao, G.~Wan, and K.~Li.
\newblock Semi-direct tracking and mapping with rgb-d camera for mav.
\newblock {\em Multimedia Tools and Applications}, 75:1--25, 2016.

\bibitem{bu2016map2dfusion}
S.~Bu, Y.~Zhao, G.~Wan, and Z.~Liu.
\newblock Map2dfusion: real-time incremental uav image mosaicing based on
  monocular slam.
\newblock In {\em Intelligent Robots and Systems (IROS), 2016 IEEE/RSJ
  International Conference on}, pages 4564--4571. IEEE, 2016.

\bibitem{burri2016euroc}
M.~Burri, J.~Nikolic, P.~Gohl, T.~Schneider, J.~Rehder, S.~Omari, M.~W.
  Achtelik, and R.~Siegwart.
\newblock The euroc micro aerial vehicle datasets.
\newblock {\em The International Journal of Robotics Research},
  35(10):1157--1163, 2016.

\bibitem{cadena2016past}
C.~Cadena, L.~Carlone, H.~Carrillo, Y.~Latif, D.~Scaramuzza, J.~Neira, I.~Reid,
  and J.~J. Leonard.
\newblock Past, present, and future of simultaneous localization and mapping:
  Toward the robust-perception age.
\newblock {\em IEEE Transactions on Robotics}, 32(6):1309--1332, 2016.

\bibitem{cummins2008fabmap}
M.~Cummins and P.~Newman.
\newblock Fab-map: Probabilistic localization and mapping in the space of
  appearance.
\newblock {\em The International Journal of Robotics Research}, 27(6):647--665,
  2008.

\bibitem{davison2007monoslam}
A.~J. Davison, I.~D. Reid, N.~D. Molton, and O.~Stasse.
\newblock Monoslam: Real-time single camera slam.
\newblock {\em IEEE transactions on pattern analysis and machine intelligence},
  29(6):1052--1067, 2007.

\bibitem{dellaert2012gtsam}
F.~Dellaert.
\newblock Factor graphs and gtsam: A hands-on introduction.
\newblock Technical report, Georgia Institute of Technology, 2012.

\bibitem{durrant2006simultaneous}
H.~Durrant-Whyte and T.~Bailey.
\newblock Simultaneous localization and mapping: part i.
\newblock {\em IEEE robotics \& automation magazine}, 13(2):99--110, 2006.

\bibitem{engel2018dso}
J.~Engel, V.~Koltun, and D.~Cremers.
\newblock Direct sparse odometry.
\newblock {\em IEEE transactions on pattern analysis and machine intelligence},
  40(3):611--625, 2018.

\bibitem{engel2014lsd}
J.~Engel, T.~Sch{\"o}ps, and D.~Cremers.
\newblock Lsd-slam: Large-scale direct monocular slam.
\newblock In {\em Computer Vision--ECCV 2014}, pages 834--849. Springer, 2014.

\bibitem{engel2015sterolsd}
J.~Engel, J.~Stuckler, and D.~Cremers.
\newblock Large-scale direct slam with stereo cameras.
\newblock In {\em Intelligent Robots and Systems (IROS), 2015 IEEE/RSJ
  International Conference on}, pages 1935--1942. IEEE, 2015.

\bibitem{engel2016photometrically}
J.~Engel, V.~Usenko, and D.~Cremers.
\newblock A photometrically calibrated benchmark for monocular visual odometry.
\newblock {\em arXiv preprint arXiv:1607.02555}, 2016.

\bibitem{fischler1981random}
M.~A. Fischler and R.~C. Bolles.
\newblock Random sample consensus: a paradigm for model fitting with
  applications to image analysis and automated cartography.
\newblock {\em Communications of the ACM}, 24(6):381--395, 1981.

\bibitem{fischler2014readings}
M.~A. Fischler and O.~Firschein.
\newblock {\em Readings in computer vision: issues, problem, principles, and
  paradigms}.
\newblock Elsevier, 2014.

\bibitem{forster2017manifold}
C.~Forster, L.~Carlone, F.~Dellaert, and D.~Scaramuzza.
\newblock On-manifold preintegration for real-time visual--inertial odometry.
\newblock {\em IEEE Transactions on Robotics}, 33(1):1--21, 2017.

\bibitem{forster2014svo}
C.~Forster, M.~Pizzoli, and D.~Scaramuzza.
\newblock Svo: Fast semi-direct monocular visual odometry.
\newblock In {\em Robotics and Automation (ICRA), 2014 IEEE International
  Conference on}, pages 15--22. IEEE, 2014.

\bibitem{forster2017svo}
C.~Forster, Z.~Zhang, M.~Gassner, M.~Werlberger, and D.~Scaramuzza.
\newblock Svo: Semidirect visual odometry for monocular and multicamera
  systems.
\newblock {\em IEEE Transactions on Robotics}, 33(2):249--265, 2017.

\bibitem{fryer1986lens}
J.~G. Fryer and D.~C. Brown.
\newblock Lens distortion for close-range photogrammetry.
\newblock {\em Photogrammetric engineering and remote sensing}, 52(1):51--58,
  1986.

\bibitem{galvez2012dbow2}
D.~G\'alvez-L\'opez and J.~D. Tard\'os.
\newblock Bags of binary words for fast place recognition in image sequences.
\newblock {\em IEEE Transactions on Robotics}, 28(5):1188--1197, October 2012.

\bibitem{gao2018ldso}
X.~Gao, R.~Wang, N.~Demmel, and D.~Cremers".
\newblock Ldso: Direct sparse odometry with loop closure".
\newblock In {\em International Conference on Intelligent Robots and Systems
  (IROS)}, October 2018.

\bibitem{gao2003complete}
X.-S. Gao, X.-R. Hou, J.~Tang, and H.-F. Cheng.
\newblock Complete solution classification for the perspective-three-point
  problem.
\newblock {\em IEEE transactions on pattern analysis and machine intelligence},
  25(8):930--943, 2003.

\bibitem{KITTI_dataset}
A.~Geiger, P.~Lenz, and R.~Urtasun.
\newblock Are we ready for autonomous driving? the kitti vision benchmark
  suite.
\newblock In {\em IEEE Conference on Computer Vision and Pattern Recognition
  (CVPR)}, 2012.

\bibitem{geiger2012we}
A.~Geiger, P.~Lenz, and R.~Urtasun.
\newblock Are we ready for autonomous driving? the kitti vision benchmark
  suite.
\newblock In {\em Computer Vision and Pattern Recognition (CVPR), 2012 IEEE
  Conference on}, pages 3354--3361. IEEE, 2012.

\bibitem{glover2012fabmap}
A.~Glover, W.~Maddern, M.~Warren, S.~Reid, M.~Milford, and G.~Wyeth.
\newblock Openfabmap: An open source toolbox for appearance-based loop closure
  detection.
\newblock In {\em Robotics and Automation (ICRA), 2012 IEEE International
  Conference on}, pages 4730--4735, May 2012.

\bibitem{grisetti2011g2o}
G.~Grisetti, H.~Strasdat, K.~Konolige, and W.~Burgard.
\newblock g2o: A general framework for graph optimization.
\newblock In {\em IEEE International Conference on Robotics and Automation},
  2011.

\bibitem{handa:etal:ICRA2014}
A.~Handa, T.~Whelan, J.~McDonald, and A.~Davison.
\newblock A benchmark for {RGB-D} visual odometry, {3D} reconstruction and
  {SLAM}.
\newblock In {\em IEEE Intl. Conf. on Robotics and Automation, ICRA}, Hong
  Kong, China, May 2014.

\bibitem{handa2014benchmark}
A.~Handa, T.~Whelan, J.~McDonald, and A.~J. Davison.
\newblock A benchmark for rgb-d visual odometry, 3d reconstruction and slam.
\newblock In {\em Robotics and automation (ICRA), 2014 IEEE international
  conference on}, pages 1524--1531. IEEE, 2014.

\bibitem{hartley2003multiple}
R.~Hartley and A.~Zisserman.
\newblock {\em Multiple view geometry in computer vision}.
\newblock Cambridge university press, 2003.

\bibitem{horn1987closed}
B.~K. Horn.
\newblock Closed-form solution of absolute orientation using unit quaternions.
\newblock {\em JOSA A}, 4(4):629--642, 1987.

\bibitem{kannala2006generic}
J.~Kannala and S.~S. Brandt.
\newblock A generic camera model and calibration method for conventional,
  wide-angle, and fish-eye lenses.
\newblock {\em IEEE transactions on pattern analysis and machine intelligence},
  28(8):1335--1340, 2006.

\bibitem{kerl2013dense}
C.~Kerl, J.~Sturm, and D.~Cremers.
\newblock Dense visual slam for rgb-d cameras.
\newblock In {\em Intelligent Robots and Systems (IROS), 2013 IEEE/RSJ
  International Conference on}, pages 2100--2106. IEEE, 2013.

\bibitem{klein2007parallel}
G.~Klein and D.~Murray.
\newblock Parallel tracking and mapping for small ar workspaces.
\newblock In {\em Mixed and Augmented Reality, 2007. ISMAR 2007. 6th IEEE and
  ACM International Symposium on}, pages 225--234. IEEE, 2007.

\bibitem{kneip2011robust}
L.~Kneip, M.~Chli, and R.~Y. Siegwart.
\newblock Robust real-time visual odometry with a single camera and an imu.
\newblock In {\em Proceedings of the British Machine Vision Conference 2011}.
  British Machine Vision Association, 2011.

\bibitem{kneip2014opengv}
L.~Kneip and P.~Furgale.
\newblock Opengv: A unified and generalized approach to real-time calibrated
  geometric vision.
\newblock In {\em Robotics and Automation (ICRA), 2014 IEEE International
  Conference on}, pages 1--8. IEEE, 2014.

\bibitem{kneip2013using}
L.~Kneip, P.~Furgale, and R.~Siegwart.
\newblock Using multi-camera systems in robotics: Efficient solutions to the
  npnp problem.
\newblock In {\em Robotics and Automation (ICRA), 2013 IEEE International
  Conference on}, pages 3770--3776. IEEE, 2013.

\bibitem{kneip2011novel}
L.~Kneip, D.~Scaramuzza, and R.~Siegwart.
\newblock A novel parametrization of the perspective-three-point problem for a
  direct computation of absolute camera position and orientation.
\newblock 2011.

\bibitem{kneip2012finding}
L.~Kneip, R.~Siegwart, and M.~Pollefeys.
\newblock Finding the exact rotation between two images independently of the
  translation.
\newblock In {\em European conference on computer vision}, pages 696--709.
  Springer, 2012.

\bibitem{lepetit2009epnp}
V.~Lepetit, F.~Moreno-Noguer, and P.~Fua.
\newblock Epnp: An accurate o (n) solution to the pnp problem.
\newblock {\em International journal of computer vision}, 81(2):155, 2009.

\bibitem{leutenegger2015okvis}
S.~Leutenegger, S.~Lynen, M.~Bosse, R.~Siegwart, and P.~Furgale.
\newblock Keyframe-based visual--inertial odometry using nonlinear
  optimization.
\newblock {\em The International Journal of Robotics Research}, 34(3):314--334,
  2015.

\bibitem{liu2018cvpr}
H.~Liu, M.~Chen, G.~Zhang, H.~Bao, and Y.~Bao.
\newblock Ice-ba: Incremental, consistent and efficient bundle adjustment for
  visual-inertial slam.
\newblock In {\em The IEEE Conference on Computer Vision and Pattern
  Recognition (CVPR)}, June 2018.

\bibitem{maruyama2016exploring}
Y.~Maruyama, S.~Kato, and T.~Azumi.
\newblock Exploring the performance of ros2.
\newblock In {\em Proceedings of the 13th International Conference on Embedded
  Software}, page~5. ACM, 2016.

\bibitem{mei2011rslam}
C.~Mei, G.~Sibley, M.~Cummins, P.~Newman, and I.~Reid.
\newblock Rslam: A system for large-scale mapping in constant-time using
  stereo.
\newblock {\em International journal of computer vision}, 94(2):198--214, 2011.

\bibitem{rmsalinas2017fbow}
R.~Muñoz-Salinas.
\newblock {FBox} fast bag of words, 2017.

\bibitem{mur2015orb}
R.~Mur-Artal, J.~Montiel, and J.~D. Tardos.
\newblock Orb-slam: a versatile and accurate monocular slam system.
\newblock {\em arXiv preprint arXiv:1502.00956}, 2015.

\bibitem{mur2014fast}
R.~Mur-Artal and J.~D. Tard{\'o}s.
\newblock Fast relocalisation and loop closing in keyframe-based slam.
\newblock In {\em Robotics and Automation (ICRA), 2014 IEEE International
  Conference on}, 2014.

\bibitem{mur2017orbslam2}
R.~Mur-Artal and J.~D. Tard{\'o}s.
\newblock Orb-slam2: An open-source slam system for monocular, stereo, and
  rgb-d cameras.
\newblock {\em IEEE Transactions on Robotics}, 33(5):1255--1262, 2017.

\bibitem{nardi2015introducing}
L.~Nardi, B.~Bodin, M.~Z. Zia, J.~Mawer, A.~Nisbet, P.~H. Kelly, A.~J. Davison,
  M.~Luj{\'a}n, M.~F. O'Boyle, G.~Riley, et~al.
\newblock Introducing slambench, a performance and accuracy benchmarking
  methodology for slam.
\newblock In {\em Robotics and Automation (ICRA), 2015 IEEE International
  Conference on}, pages 5783--5790. IEEE, 2015.

\bibitem{Newcombe2011}
R.~A. Newcombe, S.~J. Lovegrove, and A.~J. Davison.
\newblock Dtam: Dense tracking and mapping in real-time.
\newblock In {\em Computer Vision (ICCV), 2011 IEEE International Conference
  on}, pages 2320--2327. IEEE, 2011.

\bibitem{nister2004efficient}
D.~Nist{\'e}r.
\newblock An efficient solution to the five-point relative pose problem.
\newblock {\em IEEE transactions on pattern analysis and machine intelligence},
  26(6):756--770, 2004.

\bibitem{oliveira2017topometric}
G.~L. Oliveira, N.~Radwan, W.~Burgard, and T.~Brox.
\newblock Topometric localization with deep learning.
\newblock {\em arXiv preprint arXiv:1706.08775}, 2017.

\bibitem{qin2018vins}
T.~Qin, P.~Li, and S.~Shen.
\newblock Vins-mono: A robust and versatile monocular visual-inertial state
  estimator.
\newblock {\em IEEE Transactions on Robotics}, 34(4):1004--1020, 2018.

\bibitem{QUIGLEY2009ROS}
M.~Quigley, B.~Gerkey, K.~Conley, J.~Faust, T.~Foote, J.~Leibs, E.~Berger,
  R.~Wheeler, and N.~Andrew.
\newblock Ros : an open-source robot operating system.
\newblock {\em Proc. IEEE ICRA Workshop on Open Source Robotics}, 2009.

\bibitem{ORB_2011}
E.~Rublee, V.~Rabaud, K.~Konolige, and G.~Bradski.
\newblock Orb: an efficient alternative to sift or surf.
\newblock In {\em Computer Vision (ICCV), 2011 IEEE International Conference
  on}, pages 2564--2571. IEEE, 2011.

\bibitem{scaramuzza2006flexible}
D.~Scaramuzza, A.~Martinelli, and R.~Siegwart.
\newblock A flexible technique for accurate omnidirectional camera calibration
  and structure from motion.
\newblock In {\em Computer Vision Systems, 2006 ICVS'06. IEEE International
  Conference on}, pages 45--45. IEEE, 2006.

\bibitem{schubert2018tum}
D.~Schubert, T.~Goll, N.~Demmel, V.~Usenko, J.~Stueckler, and D.~Cremers.
\newblock The tum vi benchmark for evaluating visual-inertial odometry.
\newblock In {\em International Conference on Intelligent Robots and Systems
  (IROS)}, October 2018.

\bibitem{selig2004lie}
J.~Selig.
\newblock Lie groups and lie algebras in robotics.
\newblock In {\em Computational Noncommutative Algebra and Applications}, pages
  101--125. Springer, 2004.

\bibitem{stewenius2006recent}
H.~Stewenius, C.~Engels, and D.~Nist{\'e}r.
\newblock Recent developments on direct relative orientation.
\newblock {\em ISPRS Journal of Photogrammetry and Remote Sensing},
  60(4):284--294, 2006.

\bibitem{TUM2012STYRM}
J.~Sturm, N.~Engelhard, F.~Endres, W.~Burgard, and D.~Cremers.
\newblock A benchmark for the evaluation of rgb-d slam systems.
\newblock In {\em IEEE/RSJ International Conference on Intelligent Robots and
  Systems}, pages 573--580, Oct 2012.

\bibitem{sturm2012benchmark}
J.~Sturm, N.~Engelhard, F.~Endres, W.~Burgard, and D.~Cremers.
\newblock A benchmark for the evaluation of rgb-d slam systems.
\newblock In {\em Intelligent Robots and Systems (IROS), 2012 IEEE/RSJ
  International Conference on}, pages 573--580. IEEE, 2012.

\bibitem{sweeney2015theia}
C.~Sweeney.
\newblock Theia multiview geometry library: Tutorial \& reference.
\newblock {\em University of California Santa Barbara}, 2, 2015.

\bibitem{urban2016multicol}
S.~Urban and S.~Hinz.
\newblock Multicol-slam-a modular real-time multi-camera slam system.
\newblock {\em arXiv preprint arXiv:1610.07336}, 2016.

\bibitem{von2018vidso}
L.~von Stumberg, V.~Usenko, and D.~Cremers.
\newblock Direct sparse visual-inertial odometry using dynamic marginalization.
\newblock {\em arXiv preprint arXiv:1804.05625}, 2018.

\bibitem{wang2017deepvo}
S.~Wang, R.~Clark, H.~Wen, and N.~Trigoni.
\newblock Deepvo: Towards end-to-end visual odometry with deep recurrent
  convolutional neural networks.
\newblock In {\em Robotics and Automation (ICRA), 2017 IEEE International
  Conference on}, pages 2043--2050. IEEE, 2017.

\bibitem{whelan2016elasticfusion}
T.~Whelan, R.~F. Salas-Moreno, B.~Glocker, A.~J. Davison, and S.~Leutenegger.
\newblock Elasticfusion: Real-time dense slam and light source estimation.
\newblock {\em The International Journal of Robotics Research},
  35(14):1697--1716, 2016.

\bibitem{wu2011siftgpu}
C.~Wu.
\newblock Siftgpu: A gpu implementation of scale invariant feature transform
  (sift)(2007).
\newblock {\em URL http://cs. unc. edu/\~{} ccwu/siftgpu}, 2011.

\bibitem{wu2011pba}
C.~Wu, S.~Agarwal, B.~Curless, and S.~M. Seitz.
\newblock Multicore bundle adjustment.
\newblock In {\em Computer Vision and Pattern Recognition (CVPR), 2011 IEEE
  Conference on}, pages 3057--3064. IEEE, 2011.

\bibitem{yin2018geonet}
Z.~Yin and J.~Shi.
\newblock Geonet: Unsupervised learning of dense depth, optical flow and camera
  pose.
\newblock In {\em Proceedings of the IEEE Conference on Computer Vision and
  Pattern Recognition (CVPR)}, volume~2, 2018.

\bibitem{zhou2017unsupervised}
T.~Zhou, M.~Brown, N.~Snavely, and D.~G. Lowe.
\newblock Unsupervised learning of depth and ego-motion from video.
\newblock In {\em CVPR}, volume~2, page~7, 2017.

\end{thebibliography}
}

\end{document}